\documentclass[journal]{IEEEtran}

\usepackage{cite}

\ifCLASSINFOpdf
\else
\fi
\usepackage{array}

 \usepackage[caption=false,font=footnotesize]{subfig}

\usepackage{times}
\usepackage{epsfig}
\usepackage{graphicx}
\usepackage{amsmath}
\usepackage{amssymb}
\usepackage{epstopdf}
\usepackage{multirow}
\newcommand{\tabincell}[2]{\begin{tabular}{@{}#1@{}}#2
\end{tabular}}
\usepackage{subfig}

\usepackage{color}

\begin{document}
%
\title{Depth Pooling Based Large-scale 3D Action Recognition with Convolutional Neural Networks}
%
%
%

\author{Pichao Wang,~\IEEEmembership{Member,~IEEE,}
        Wanqing Li,~\IEEEmembership{Senior Member,~IEEE,}\\
        Zhimin Gao,
        Chang Tang, and 
        Philip Ogunbona,~\IEEEmembership{Senior Member,~IEEE,}
\thanks{Manuscript received on 15-Apr-2017; revised on 10-Dec-2017 and 25-Feb-2018; accepted on 19-Mar-2018  (Corresponding author: Zhimin Gao)}
\thanks{P. Wang, W. Li, Z. Gao and P. Ogunbona are with the 
Advanced Multimedia Research Lab, University of Wollongong, Wollongong, 
Australia.
(e-mail: pw212@uowmail.edu.au; wanqing@uow.edu.au; zg126@uowmail.edu.au;philipo@uow.edu.au).}
\thanks{C. Tang are with the School of Computer Science, China University of Geosciences, Wuhan 430074, PR China (e-mail:happytangchang@gmail.com).}}

%
%

\markboth{IEEE TRANSACTIONS ON MULTIMEDIA,~Vol.~X, No.~X, XX~2018 (WILL BE INSERTED BY THE EDITOR)}%
{Shell \MakeLowercase{\textit{et al.}}:Depth Pooling Based Large-scale 3D Action Recognition with Convolutional Neural Networks}
%



\maketitle

\begin{abstract}
This paper proposes three simple, compact yet effective representations of depth sequences, referred to respectively as Dynamic Depth Images (DDI), Dynamic Depth Normal Images (DDNI) and Dynamic Depth Motion Normal Images (DDMNI), for both  isolated and continuous action recognition. These dynamic images are constructed from a segmented sequence of depth maps using hierarchical bidirectional rank pooling to effectively capture the spatial-temporal information. Specifically, DDI exploits the dynamics of postures over time and DDNI and DDMNI exploit the 3D structural information captured by depth maps. Upon the proposed representations, a ConvNet based method is developed for action recognition. The image-based representations enable us to fine-tune the existing Convolutional Neural Network (ConvNet) models trained on image data without training a large number of parameters from scratch. The proposed method achieved the state-of-art results on three large datasets, namely, the Large-scale Continuous Gesture Recognition Dataset (means Jaccard index $0.4109$), the Large-scale Isolated Gesture Recognition Dataset ($59.21\%$), and the NTU RGB+D Dataset ($87.08\%$ cross-subject and $84.22\%$ cross-view) even though only the depth modality was used.

\end{abstract}

\begin{IEEEkeywords}
Large-scale, Depth, Action Recognition, Convolutional Neural Networks.
\end{IEEEkeywords}

%
\IEEEpeerreviewmaketitle

\section{Introduction}
\IEEEPARstart{M}{icrosoft}  Kinect Sensors 
provide an affordable technology to capture depth maps and RGB images in 
real-time. Recognition of human actions from depth data is one of the most active research topics in multimedia signal processing. Compared to traditional images, depth maps offer better geometric 
cues and less sensitivity to illumination changes for action recognition. Since the first work of such a type~\cite{li2010action} reported in 2010, many methods~\cite{aggarwal2014human,presti20163d,zhang2016rgb} have been proposed based on specific hand-crafted feature descriptors extracted from depth. Most previous action recognition methods primarily focus on modeling spatial configuration based on hand-designed features and adopt dynamic time warpings (DTWs), Fourier temporal pyramid (FTP) or hidden Markov models (HMMs) to represent temporal information. However, these hand-crafted features are always shallow and dataset-dependent. Recently, recurrent neural networks (RNNs)~\cite{du2015hierarchical,veeriah2015differential,liu2016spatio} have also been adopted for action recognition. RNNs tend to overemphasize the temporal information especially when there is insufficient training data, leading to overfitting. Up to now, it remains unclear how video could be effectively represented and fed to deep neural networks for classification.  For instance, a video can be considered as a sequence of still images~\cite{yue2015beyond} with some form of temporal
smoothness~\cite{simonyan2014two,jayaraman2016slow} and fed into a convolutional network (ConvNet). It is also possible to extend the ConvNet to a third temporal dimension~\cite{ji20133d,tran2015learning} by replacing 2D filters with their 3D equivalent. Another possibility is to regard the video as the output of a neural network encoder~\cite{srivastava2015unsupervised}. Lastly, one can treat the video as a sequence of images that can be fed to a
RNN~\cite{donahue2015long,veeriah2015differential,shahroudy2016ntu}. 

Inspired by the promising performance of recently introduced rank pooling machine~\cite{Fernando2015a,bilen2016dynamic,Fernando2016a} on RGB videos, we propose to extend the rank pooling method to encode depth map sequences into dynamic images. However, due to the insensitivity of depth to motion, the generated dynamic images by directly applying rank pooling method~\cite{bilen2016dynamic} are not enough to exploit rich spatial-temporal and structural information. In this paper, to take full use of depth advantages, namely, being insensitive to illumination changes and providing the 3D surface information of the objects, three image-based depth representations, referred to as respectively Dynamic Depth Image (DDI), Dynamic Depth Normal Image (DDNI) and Dynamic Depth Motion Normal Image (DDMNI), are proposed and they are constructed by applying rank pooling hierarchically and bidirectionally to a segmented sequence of depth maps. The hierarchical bidirectional rank pooling overcomes the drawbacks of the conventional ranking method~\cite{bilen2016dynamic}, that is, it eliminates the bias to the past frames in the conventional rank pooling and exploits both the higher order and non-linear dynamics of depth data simultaneously. Specifically, DDIs mainly exploit the dynamic of postures, and DDNI and DDMNI, built upon norm vectors, effectively exploit the 3D structural information captured by depth maps. Such representations make it possible to use a standard ConvNet architecture to learn discriminative ``dynamic" features and to tune ConvNet models trained from image data on small annotated depth data without training the parameters of the ConvNet from scratch. For instance, the large-scale isolated gesture recognition challenge~\cite{wanchalearn} has on average only 144 video clips per class compared to 1200 images per class in ImageNet.  Experimental results have shown that the three representations can improve the accuracy of action recognition substantially compared to the state-of-the-art methods. 

Part of the work~\cite{pichaoicprwb,pichaoicprwa} was reported in the ChaLearn Looking at People (LAP) challenge \footnote{http://gesture.chalearn.org/icpr2016\_contest}\footnote{The team won the second place in the Isolated Gesture Recognition, third place in the Continuous Gesture Recognition, and two best paper awards} in conjunction with ICPR 2016. The key novelty of the method is to encode the geometric, motion and structural information of human actions into three dynamic depth images though ranking pooling. Compared to the two workshop papers \cite{pichaoicprwb,pichaoicprwa}, the extension includes (1) the bidirectional rank pooling in the method of \cite{pichaoicprwb} is replaced with {\em hierarchical} and {\em bidirectional} rank pooling to capture both high order and non-linear dynamics more effectively; (2) The extended method is applied to continuous action recognition; (3) the method is also evaluated on the large and challenging NTU RGB-D dataset in addition to the ChaLearn LAP datasets and state-of-the-art results are achieved on the three large datasets using depth modality only; and (4) more analysis are presented in this paper.



The rest of this paper is organized as follows. Section~\ref{sec:related_work} briefly reviews the related works. Details of the proposed method are described in Section~\ref{sec:proposed_method}. Experimental results and analysis are presented in Section~\ref{sec:experiments}. Section~\ref{sec:conclusions} concludes the paper.

\begin{figure*}[t]
\begin{center}
{\includegraphics[height = 90mm, width = 185mm]{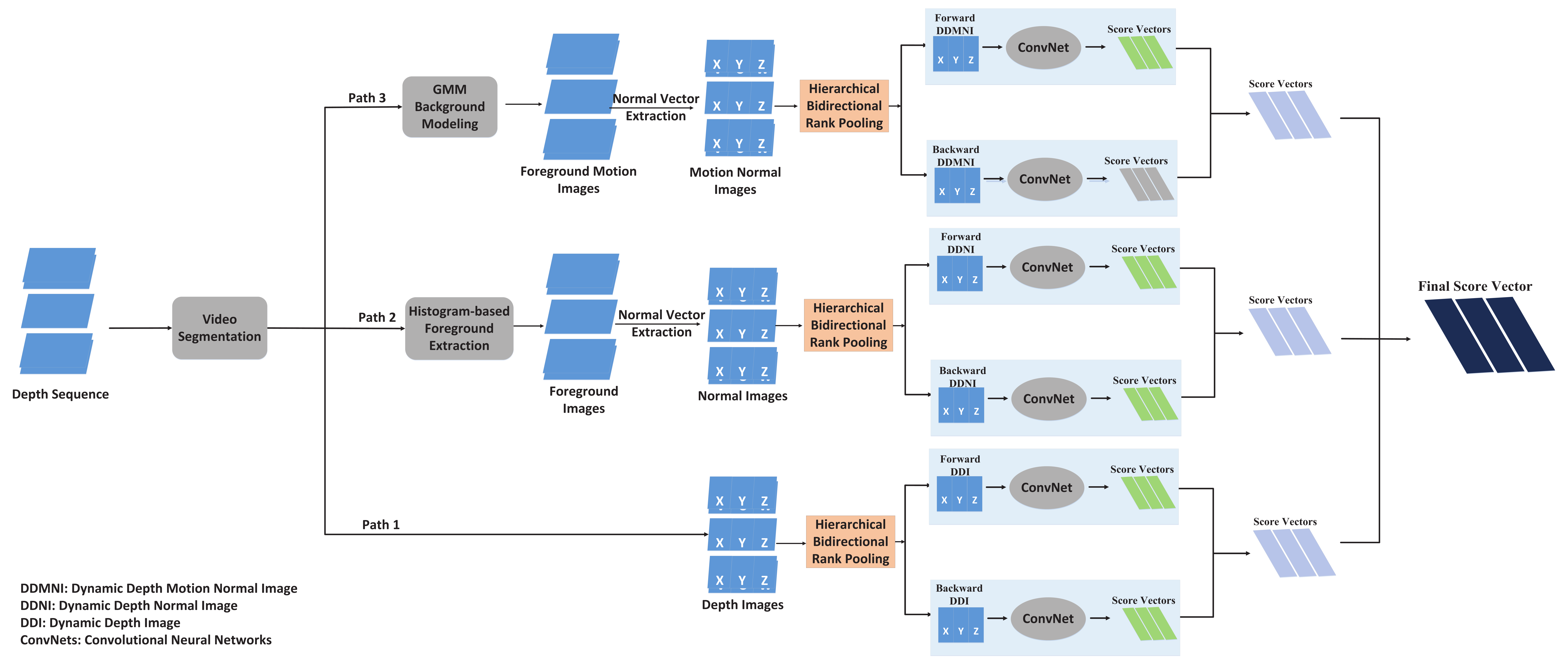}}
\end{center}
\caption{The framework of the proposed method.}
\label{fig:framework}
\end{figure*}

\section{Related Work}
\label{sec:related_work}
In this section, recent works related to the paper, including action segmentation, action recognition based on depth and deep learning-based action recognition, are briefly reviewed. Comprehensive reviews can be found in ~\cite{d2016recent,pisharady2015recent,escalera2016challenges,cheng2016,herath2016going}.

\subsection{Action Segmentation}
Most existing works on action recognition reported to date focus on the classification of individual actions by assuming that instances of individual actions have been isolated or segmented from a video or a stream of depth maps before the classification. In the case of continuous recognition, the input stream usually contains unknown numbers, unknown orders and unknown boundaries of actions and both segmentation and recognition have to be solved at the same time. There are three common approaches to such continuous recognition. The first approach is to tackle the segmentation and classification of actions separately and sequentially. The key advantages of this approach are that different features can be used for segmentation and classification and existing classification methods can be leveraged. The disadvantages are that both segmentation and classification could be the bottleneck of the systems and they can hardly be optimized together.  The second approach is to apply classification to a sliding temporal window and aggregate the window based classification to achieve final segmentation and classification. One of the key challenges in this approach is the difficulty of setting the size of sliding window because durations of different gestures or actions can vary significantly.  The third approach is to perform the segmentation and classification simultaneously. 
With respect to action segmentation, the popular and widely used method employs dynamic time warping~(DTW) to determine the delimiting frames of individual actions~\cite{jmlr14jun, tip14-jun, paa15-jun}.  Difference images are first obtained by subtracting two consecutive grey-scale images and each difference image is partitioned into a grid of $3\times3$ cells. Each cell is then represented by the average value of pixels within this cell. The matrix of the cells in a difference image is flattened as a vector called motion feature and calculated for each frame in the video, excluding the final frame. This
results in a $9\times{}(K-1)$ matrix of motion features for a video with $K$ frames. The motion feature matrix is extracted from both test video and training video which consists of multiple actions. The two matrices are treated as two temporal sequences with each motion feature as a feature vector at an instant of time. The distance between two feature vectors is defined as the negative Euclidean distance and a matrix containing DTW  distances (measuring similarity between two temporal sequences) between the two sequences is then calculated and analyzed by Viterbi algorithm~\cite{pieee73forney} to segment the actions. 
Another category of action segmentation methods from a multi-action video is based on appearance. Upon the general assumption that the \textit{start} and \textit{end} frames of adjacent actions are similar, correlation coefficients~\cite{jmlr12-yuiman} and K-nearest neighbour algorithm with histogram of oriented gradient~(HOG)~\cite{cvpr12-wu} are used to identify the \textit{start} and \textit{end} frames of actions.  Jiang et al.~\cite{jiang2015multi} proposed a method based on quantity of movement~(QOM) by assuming the same \textit{start} pose among different actions. Candidate delimiting frames are chosen based on the global QOM. After a refining stage which employs a sliding window to keep the frame with minimum QOM in each windowing session, the \textit{start} and \textit{end} frames are assumed to be the remaining frames. This paper adopts the QOM based method and its details will be presented in Section~\ref{subsec:video-seg}.

\subsection{Depth Based Action Recognition}
Many methods have been reported to date on depth map-based action recognition. Li et al. \cite{li2010action} sampled points from a depth map to obtain a bag of 3D points to encode spatial information and employ an expandable graphical model to encode temporal information \cite{li2008}. Yang et al. \cite{Yang2012a} stacked differences between projected depth maps as a depth motion map (DMM) and then used HOG to extract relevant features from the DMM. This method transforms the problem of action recognition from spatio-temporal space to spatial space.  In \cite{Oreifej2013}, a feature called Histogram of Oriented 4D Normals (HON4D) was proposed; surface normal is extended to 4D space and quantized by regular polychorons. Following this method, Yang and Tian \cite{yangsuper} cluster hypersurface normals and form the polynormal which can be used to jointly capture the local motion and geometry information. Super Normal Vector (SNV) is generated by aggregating the low-level polynormals. In \cite{lurange}, a fast binary range-sample feature was proposed based on a test statistic by carefully designing the sampling scheme to exclude most pixels that fall into the background and to incorporate spatio-temporal cues. All of previous works are based on hand-crafted features which is shallow and dataset-dependent. With the development of deep learning, Wang et al.~\cite{pichao2015,pichaoTHMS} applied ConvNets to depth action recognition based on the variants of DMM~\cite{Yang2012a}, which is sensitive to noise and cannot work well in fine-grained action recognition. Wu. et al.~\cite{diwu2016} adopted a 3D ConvNet to extract features from depth data, which requires a large amount of training data to achieve the best performance.  Wang et al.~\cite{wang2017structured} and Hou et al.~\cite{Hou2018} proposed the concept of structured images for depth based action recognition, but the proposed methods need the help of skeleton data to locate the position of human body, parts and joints.

\subsection{Deep Leaning Based Motion Recognition}
Existing deep learning approaches can generally be divided into four categories based on how the video is represented and fed to a deep neural network. The first category views a video either as a set of still images~\cite{yue2015beyond} or as a short and smooth transition between similar frames~\cite{simonyan2014two}, and each color channel of the images is fed to one channel of a ConvNet. Although obviously suboptimal, considering the video as a bag of static frames performs reasonably well. The second category is to represent a video as a volume and extends ConvNet to a third, temporal dimension~\cite{ji20133d,tran2015learning} replacing 2D filters with their 3D equivalents. So far, this approach has produced little benefit, probably due to the lack of annotated training data. The third category is to treat a video as a sequence of images and feed the sequence to a RNN~\cite{donahue2015long,du2015hierarchical,veeriah2015differential,liu2016spatio}. A RNN is typically considered as memory cells, which are sensitive to both short as well as long term patterns. It parses the video frames sequentially and encode the frame-level information in the memory. However, using RNNs did not give an improvement over temporal pooling of convolutional features~\cite{yue2015beyond} or over hand-crafted features. The last category is to represent a video in one or multiple compact images and adopt available trained ConvNet architectures for fine-tuning~\cite{pichao2015,pichaoTHMS,pichao2016,bilen2016dynamic,pichaocsvt2016,wang2017cooperative}. This category has achieved state-of-the-art results of action recognition on many RGB and depth/skeleton datasets. The proposed method in this paper falls into the last category.

\section{Proposed Method}
\label{sec:proposed_method}
The proposed method consists of four stages: action segmentation, construction of the three sets of dynamic images, ConvNets training and score fusion for classification. The framework is illustrated in Fig.~\ref{fig:framework}. Given a sequence of depth maps consisting of multiple actions, the \textit{start} and \textit{end} frames of each action are identified based on quantity of movement (QOM)~\cite{jiang2015multi}. Then, three sets of dynamic images are constructed for each action segment and used as the input to six ConvNets for multiple score fusion-based classification. Details are presented in the rest of this section.

\subsection{Action Segmentation}\label{subsec:video-seg}
Previous works on action recognition mainly focus on the classification of segmented actions. In the case of continuous recognition, both segmentation and recognition have to be solved. This paper tackles the segmentation and classification of actions separately and sequentially. 

\begin{figure}[t]
\begin{center}
{\includegraphics[height = 50mm, width = 90mm]{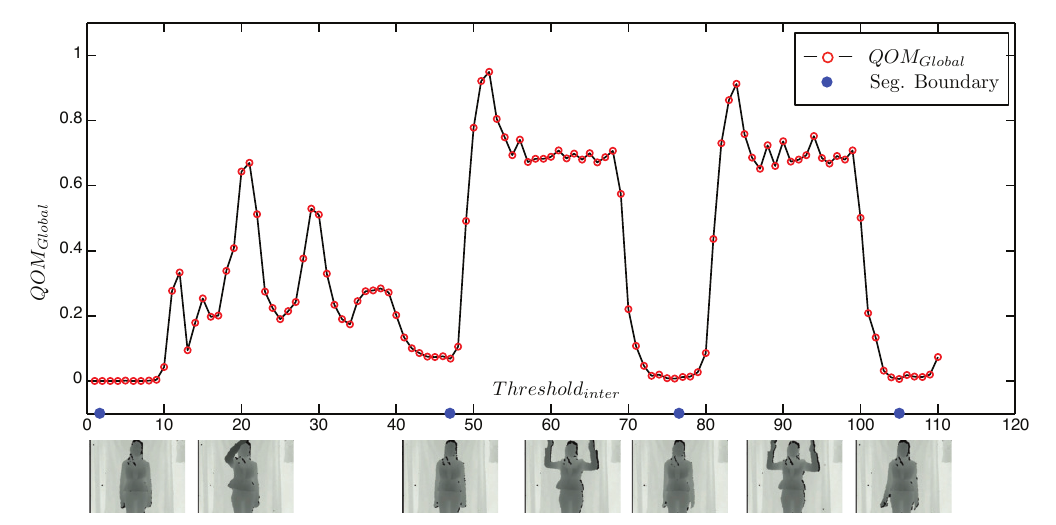}}
\end{center}
\caption{An example of illustrating the inter-action segmentation results. Figure from~\cite{jiang2015multi}.}
\label{segmentation}
\end{figure}

Given a  sequence of depth maps that contains multiple actions, each frame has the relevant movement with respect to its adjacent frame and the first frame. The \textit{start} and \textit{end} frames of each action is detected based on quantity of movement~(QOM)~\cite{jiang2015multi} by assuming that all actions starts from a similar pose.  For a multi-action depth sequence $I$, the QOM for frame $t$ is defined as
\begin{equation}
\begin{aligned}
QOM_{Global}(I,t) = \sum_{m,n}\psi(I_{t}(m,n), I_{1}(m,n))
\end{aligned},
\end{equation}
where $(m,n)$ is the pixel location and the indicator function $\psi(x,y)$ is defined as 
$$\psi(x,y) =
\begin{cases}
1& \text{$if |x - y| \geqslant Threshold_{QOM}$};\\
0& \text{otherwise}
\end{cases}$$
$Threshold_{QOM}=60$ is a threshold predefined emperically. A set of frame indices of 
candidate delimiting frames is initialized by choosing frames with lower global QOMs than a $threshold_{inter}$. The $threshold_{inter}$ is calculated by adding the mean to twice the standard deviation of the global QOMs extracted from first and last $12.5\%$ of the average action sequence length $L$ calculated from the training actions. A sliding window of size $\frac{L}{2}$ is then used to refine the candidate set.  In each windowing session only the  frame with a minimum global QOM is considered to be the boundary frame of two consecutive actions. After the refinement, the remaining frames are expected to be the delimiting frames of actions, as shown in Fig.~\ref{segmentation}.

\subsection{Construction of Dynamic Images}

The three sets of dynamic images, Dynamic Depth Images (DDIs), Dynamic Depth Normal Images (DDNIs) and Dynamic Depth Motion Normal Images (DDMNIs) are constructed from a segmented sequence of depth maps through hierarchical bidirectional rank pooling. They aim to exploit shape, motion and structural information captured by a depth sequence at different spatial and temporal scales. To this end, the conventional ranking pooling~\cite{bilen2016dynamic} is extended to the hierarchical bidirectional rank pooling.

The conventional rank pooling~\cite{bilen2016dynamic} aggregates spatio-temporal information from one video sequence into one dynamic image. It defines a function that maps a video clip into one feature vector~\cite{bilen2016dynamic}. A \textit{rank pooling function} is formally defined as follows.
 
Rank Pooling. Let a depth map sequence with $k$ frames be represented as  $<d_{1},d_{2},...,d_{t},...,d_{k}>$, where $d_{t}$ is the average of depth features over the frames up to $t$-timestamp.  At each time $t$, a score $r_{t} = \omega^{T}\cdot d_{t}$ is assigned. The score satisfies $r_{i} > r_{j} \Longleftrightarrow i > j$. In general, more recent frames are associated with larger scores. The process of rank pooling is to find $\omega^{*}$ that satisfies the following objective function:

\begin{equation}
\begin{aligned}
\ \mathop{\arg\min}_{\omega} \dfrac{1}{2}\parallel \omega \parallel^{2} + \lambda \sum\limits_{i > j} \xi_{ij}\\ s.t.~~ \omega^{T}\cdot(d_{i}-d_{j})\geq 1 -\xi_{ij}, \xi_{ij} \geq 0\\
\end{aligned},
\end{equation}
where $\xi_{ij}$ is a slack variable. Since the score $r_i$ assigned to frame $i$ is often defined as the order of the frame in the sequence, $\omega^{*}$ aggregates information from all of the frames in the sequence and can be used as a descriptor of the sequence. In this paper, the rank pooling is directly applied on the pixels of depth maps and the $\omega^{*}$ is of the same size as depth maps and forms a dynamic depth image (DDI).

However, the conventional ranking pooling method has two drawbacks.  Firstly, it treats a video sequence in a single temporal scale which is usually too shallow~\cite{Fernando2016a}. Secondly, since in rank pooling the averaged feature up to time t is used to classify frame t, the pooled feature is biased towards beginning frames of a depth sequence, hence, frames at the beginning has more influence to $\omega^{*}$. This is not justifiable in action recognition as there is no prior knowledge on which frames are more important than other frames.

To overcome the first drawback, it is proposed that the ranking pooling is applied recursively to sliding windows over several \textit{rank pooling layer}. This recursive process can effectively explore the high-order and non-linear dynamics of a depth sequence. The \textit{rank pooling layer} is defined as follows:


Definition 2 (Rank Pooling Layer).
Let ${I^{(l)}} = \left\langle {i_1^{(l)},...,i_n^{(l)}} \right\rangle $ denote the input sequence/subsequence that contains $n$ frames; ${{M_l}}$ is the window size; and ${{S_l}}$ is a stride in the $l_{th}$ layer.  The subsequences of ${I^{(l)}}$ can be defined as $ I_t^{(l)} = \left\langle {i_t^{(l)},...,i_{t + {M_l} - 1}^{(l)}} \right\rangle $, where $t \in \left\{ {1,\;{S_l} + 1,\;2{S_l} + 1, \ldots } \right\}$. By applying the \textit{rank pooling function} on the subsequences respectively, the outputs of $l_{th}$ layer constitute the ${(l + 1)_{th}}$ layer, which can be represented as ${I^{(l+1)}} = \left\langle { \ldots ,i_t^{(l + 1)},...} \right\rangle $.

\begin{figure}[t]
\begin{center}
{\includegraphics[height = 60mm, width = 85mm]{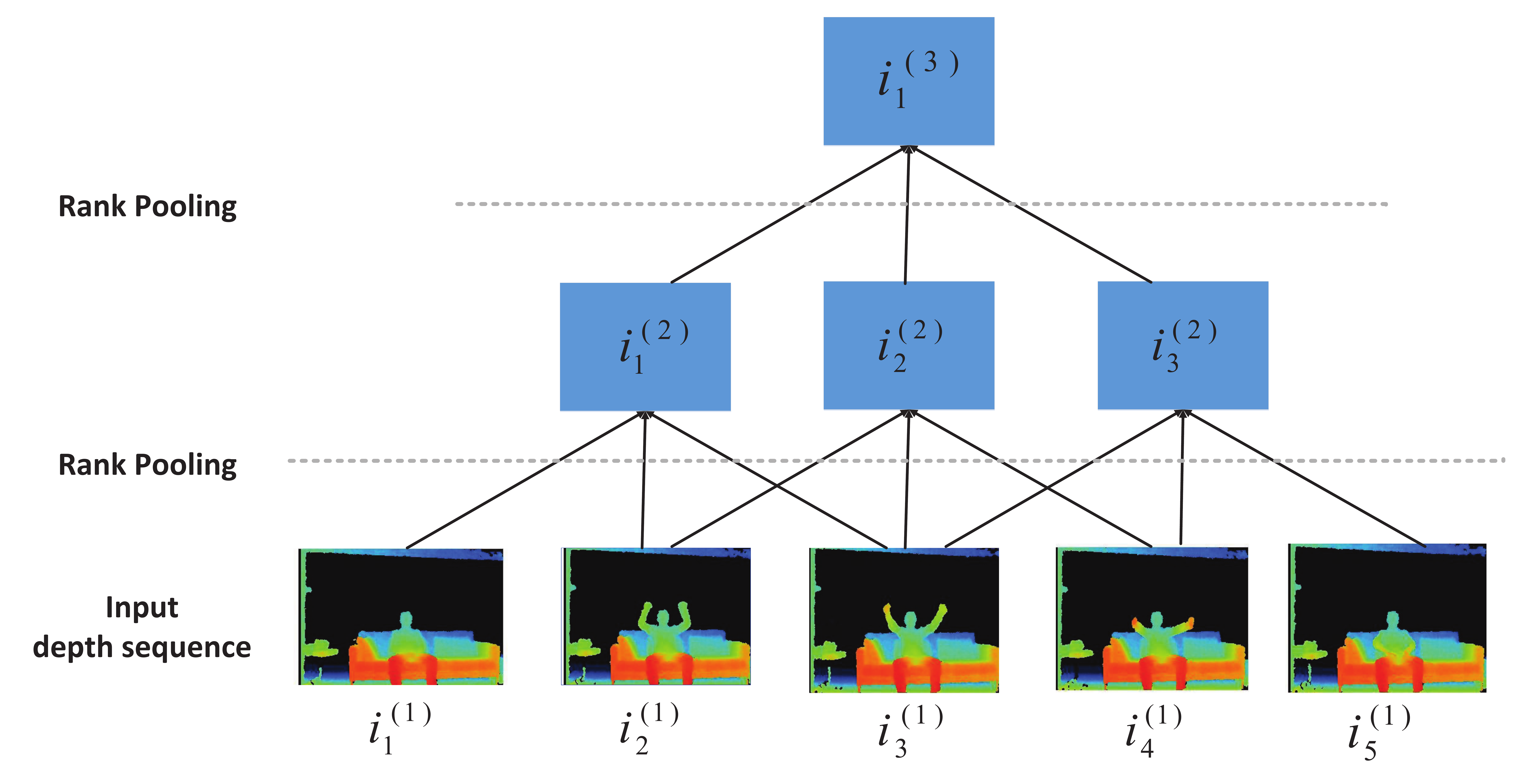}}
\end{center}
\caption{Illustration of a two layered rank pooling with window size three (${{M_l}}$ = 3) and stride one (${{S_l}}$ = 1).}
\label{temporal}
\end{figure}

 ${I^{(l)}}$ to ${I^{(l+1)}}$ forms one layer of temporal hierarchy. Multiple \textit{rank pooling layers} can be stacked together to make the pooling higher-order. In this case, each successive layer obtains the dynamics of the previous layer. Figure~\ref{temporal} shows a hierarchical rank pooling with two layers. For the first layer, the sequence is the input depth sequence, thus $l = 1$, $n=5$; for the second layer, $l = 2$, $n=3$. By adjusting the window size and stride of each layer, the hierarchical rank pooling can explore high-order and non-linear dynamics effectively.
 
 To address the second drawback, it is proposed to to apply the rank pooling bidirectionally.

Bidirectional Rank Pooling is to apply the rank pooling forward and backward to a sequence of depth maps. In the forward rank pooling, the $r_i$ is defined in the same order as the time-stamps of the frames. In the backward rank pooling, $r_i$ is defined in the reverse order of the time-stamps of the frames. When bidirectional rank pooling is applied to a sequence of depth maps, two DDIs, forward DDI and backward DDI, are generated.

By employing the hierarchical and bidirectional pooling together, the hierarchical bidirectional rank pooling exploits the dynamics of a depth sequence at different temporal scales and bidirectionally at the same time. It has been empirically observed that, for most actions with relatively short durations, two layers of bidirectional rank pooling is sufficient.

\subsubsection{Construction of DDI}
Given a segmented sequence of depth maps, the hierarchical bidirectional rank pooling method described above is employed directly on the depth pixels to generate two dynamic depth images (DDIs), forward DDI and backward DDI. Even though rank pooling method exploits the evolution of videos and aims to encode both the spatial and motion information into one image, it is likely to lose much motion information due to the insensitivity of depth pixels to motion. As shown in Fig.~\ref{fig:DIs}, DDIs effectively capture the posture information, similar to key poses. Moreover, compared with the dynamic images (DIs~\cite{bilen2016dynamic}), the DDIs are more effective, without having interfering texture on the body. 

\subsubsection{Construction of DDNI}


Depth images well represent the geometry of surfaces in the scene, and norm vectors is sensitive to motion of depth pixels. In order to simultaneously exploit the spatial and motion information in depth sequences, it is proposed to extract normals from depth maps and construct the so-called DDNIs (dynamic depth normal images). For each depth map, a surface normal $(n_x,n_y,n_z)$ is calculated at each pixel. Three channels $(N_x,N_y,N_z)$, referred to as a Depth Normal Image, are generated from the normals, where $(N_x,N_y,N_z)$ are respectively normal images of the three components $(n_x,n_y,n_z)$. The sequence of each DNI goes through hierarchical bidirectional rank pooling to generate two DDNIs, one being the forward DDNI and the other is the backward DDNI.

To minimize the interference of the background, it is assumed that the background in the histogram of depth maps occupies the last peak representing far distances. Specifically, pixels whose depth values are greater than a threshold defined by the last peak of the depth histogram minus a fixed tolerance  are considered as background and removed from the calculation of DDNIs by re-setting their depth values to zero. Through this simple process, most of the background can be removed and has much contribution to the DDNIs.  Samples of DDNIs can be seen in Fig.~\ref{fig:DIs}.

\subsubsection{Construction of DDMNI}

\begin{figure}[t]
\begin{center}
{\includegraphics[height = 90mm, width = 85mm]{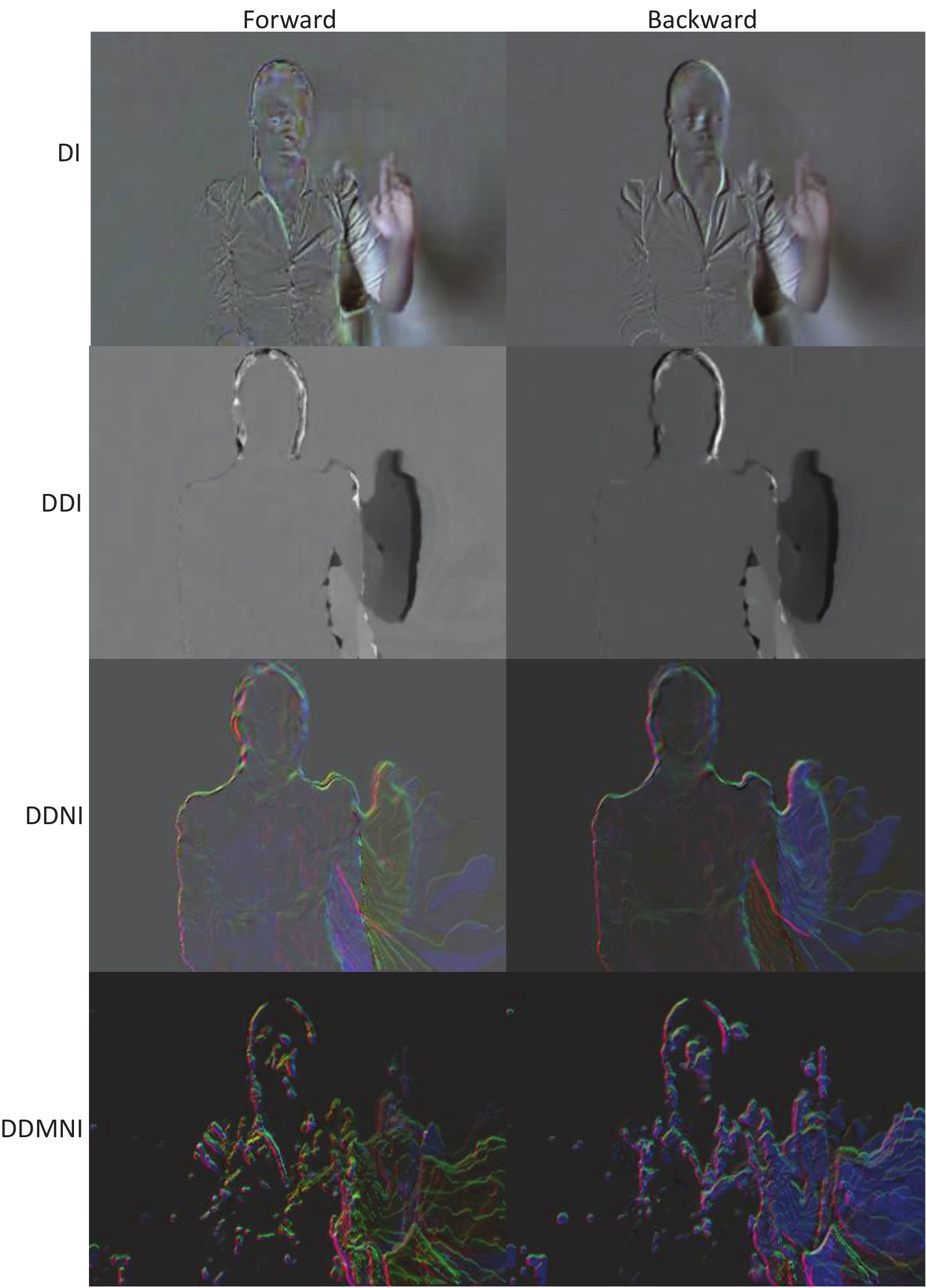}}
\end{center}
\caption{Samples of generated forward and backward DIs~\cite{bilen2016dynamic}, DDIs, DDNIs and DDMNIs for gesture Mudra1/Ardhapataka.}
\label{fig:DIs}
\end{figure}

The purpose of constructing a DDMNI is to further exploit the motion in depth maps. Gaussian mixture model (GMM) is applied to depth sequences in order to detect moving foreground. The norm vectors are extracted from the moving foreground and Depth Normal Image is constructed from the norm vectors for each depth map. Hierarchical bidirectional rank pooling is applied to the Depth Norm Image sequence, and two DDMNIs, forward DDMNI and backward DDMNI, are generated, which capture the motion information specifically well (see the illustration in Fig.~\ref{fig:DIs}).

\subsection{Network Training}
After the construction of DDIs, DDNIs and DDMNIs, there are six dynamic images, as illustrated in Fig.~\ref{fig:DIs}, for each depth map sequence. Six ConvNets were trained on the six channels individually. VGG-16~\cite{simonyan2014very} is adopted in this paper. The implementation is derived from the publicly available Caffe toolbox~\cite{jia2014caffe} based on three {NVIDIA Tesla K40 GPU} cards and one Pascal TITAN X.

The training procedure is similar to those in~\cite{simonyan2014very}. The network weights are learned using the mini-batch stochastic gradient descent with the momentum set to 0.9 and weight decay set to 0.0005. All hidden weight layers use the rectification (RELU) activation function. At each iteration, a mini-batch of 32 samples is constructed by sampling 256 shuffled training samples, and all the images are resized to 224 $\times$ 224. The learning rate is set to $10^{-3}$ for fine-tuning with pre-trained models on ILSVRC-2012, and then it is decreased according to a fixed schedule, which is kept the same for all training sets. The training undergoes 100 epochs and the learning rate decreases every 30 epochs for each ConvNet. The dropout regularization ratio is set to 0.9 to reduce complex co-adaptations of neurons in nets.

\subsection{Score Fusion for Classification}

Given a test depth video sequence (sample), three pairs of dynamic images (DDIs, DDNIs, DDMNIs) are generated and fed into six different trained ConvNets. For each image pair, multiple-score fusion was used. The score vector output from the two pair of ConvNets are multiplied in an element-wise  manner and the resultant score vectors are normalized using $L_{1}$ norm. The three normalized score vectors are then multiplied in an element-wise fashion and the max score in the resultant vector is assigned as the probability of the test sequence being the recognized class. The index of this max score corresponds to the recognized class label and expressed as follows:
\begin{equation}
label=Fin(max{(v_1\circ v_2\circ v_3\circ v_4 \circ v_5 \circ v_6)})
\end{equation}
where $v$ is a score vector, $\circ$ refers to element-wise multiplication and $Fin(\cdot)$ is a function to find the index of the element having the maximum score.

\section{Experiments}
\label{sec:experiments}
In this section, the Large-scale Isolated and Continuous Gesture Recognition datasets at the ChaLearn LAP challenge 2016 (ChaLearn LAP IsoGD Dataset and ChaLearn LAP ConGD Dataset)~\cite{ICPRW2016}, the NTU RGB+D dataset~\cite{shahroudy2016ntu}, and the corresponding evaluation protocols and results \& analysis are described. On ChaLearn LAP ConGD Dataset, action segmentation was first conducted to segment the continuous actions to isolated actions.

\subsection{Settings}
This section presents empirical studies to set the parameters for segmentation, hierarchical rank pooling, choice of networks and classifiers.  
\subsubsection{Segmentation}
A sliding window is used to refine the candidate set and in each windowing session only location of the frame with a minimum global QOM is retained as the boundary of two gestures. The window size is determined by the average action sequence length $L$ which is learned from the training samples. Theoretically, the sliding window size would affect the segmentation results. Several experiments were conducted using the DDI forward channel on the validation subset of the Chalearn LAP ConGD dataset to evaluate how the sliding window size would affect the performance. Results have shown that too large or too small sliding window size may affect adversely to the final recognition. But any value between an half or a quarter of the average gesture length worked well. Therefore, we set the window size to $\frac{L}{2}$ in all of our experiments.


\subsubsection{Parameters for hierarchical rank pooling}
For hierarchical rank pooling, generally speaking, the value of $l$ is decided by the complexity of actions/gestures in the dataset, and the $n$ is decided by the window size $M_{l}$ and stride $S_{l}$. It has been empirically observed that, for most actions with relatively short durations, two layers of bidirectional rank pooling is sufficient. To investigate the affects of these two parameters, we conducted several experiments on the ChaLearn LAP IsoGD dataset using the DDI forward channel.
The results have demonstrated that the final performance is not sensitive to the parameters ($l$, $M_{1}$, $S_{l}$). Consequently, two layered hierarchical bidirectional rank pooling method is adopted with window size ${M_l} = 3$ and stride step ${S_l} = 1$ for all the experiments reported in the paper.

\subsubsection{Choice of network architecture and classifier}
Experimental studies were also conducted on other other CNNs include AlexNet, GoogleNet, VGG-16, VGG-19 and the  new CNN architecture, MobileNetV2~\cite{sandler2018inverted}.  Results have shown that VGG-16 $(36.92\%)$ and VGG-19 $(36.96\%)$ performed better than AlexNet $(32.22\%)$, GoogleNet $(36.11\%)$ and MobileNet V2 ($34.56\%$) on the validation set. However, VGG-16 performed comparable to VGG-19. For the sake of computational cost, VGG-16 was chosen in our experiments.

Several experiments were also conducted by using kNN with $k=1,5,15,25$, linear SVM on the features from the convolutional layers of VGG-16 and the full VGG-16 on the DDI forward channel of the ChaLearn LAP IsoGD dataset. VGG-16 performed favourably to others.  

\subsection{ChaLearn LAP IsoGD Dataset}
\subsubsection{Description}
The ChaLearn LAP IsoGD Dataset is derived from the ChaLearn Gesture Dataset (CGD)~\cite{guyon2014chalearn}. It includes 47933 RGB-D depth sequences, each RGB-D video representing one gesture instance. There are 249 gestures performed by 21 different individuals.  The detailed information of this dataset are shown in Table~\ref{iso}.  In this paper, only depth maps are used to evaluate the performance of the proposed method. 

\begin{table*}[!ht]
\centering
\caption{Information of the ChaLearn LAP IsoGD Dataset. \label{iso}}
\begin{tabular}{|c|c|c|c|c|c|c|}
\hline
Sets &\# of labels &\# of gestures & \# of RGB videos & \# of depth videos & \# of subjects & label provided \\
\hline
Training & 249 & 35878 & 35878 & 35878 & 17 & Yes \\
\hline
Validation & 249 & 5784 & 5784 & 5784 & 2 & No \\
\hline
Testing & 249 &  6271 & 6271 & 6271 & 2 & No \\
\hline
All & 249 & 47933 & 47933 & 47933 & 21 & - \\
\hline
\end{tabular}
\end{table*}


\subsubsection{Evaluation Protocol}

The dataset is divided into training, validation and test sets. All three sets consist of samples of different subjects in order to ensure that the gestures of one subject in validation and test sets do not appear in the training set. 

For the isolated gesture recognition challenge, recognition rate $r$ is used as the evaluation criteria. The recognition rate is calculated as:
\begin{equation}
r = \dfrac{1}{n}\delta(p_{l}(i),t_{l}(i))
\end{equation}
where $n$ is the number of samples; $p_{l}$ is the predicted label; $t_{l}$ is the ground truth; $\delta(j_{1},j_{2}) = 1$, if $j_{1} = j_{2}$, otherwise $\delta(j_{1},j_{2}) = 0$.

\subsubsection{Experimental Results}
Table~\ref{comlimentary} shows the results of each channel. From the results we can see that DDIs achieved much better results than DDNIs and DDMNIs, and the reasons are as follows: first, the depth values are not the real depth, but they are normalized to [0,255], which distort the true 3D structure information and  affects the norm vectors extraction; second, for storage benefit, the videos are compressed at a loss level, which leads to lots of compression blocking artifacts, which makes the extraction of moving foreground and norm vectors very noisy. Even though, the three kinds of dynamic images still provide complimentary information to each other. In addition, it can be seen that the bidirectional rank pooling exploits more useful information compared to one-way rank pooling~\cite{bilen2016dynamic}, and by adopting multiply score fusion method, the accuracy is largely improved. Moreover, hierarchical rank pooling encodes the dynamic of depth sequences better compared with the conventional rank pooling method. 

\begin{table}[!ht]
\centering
\caption{Comparative Accuracy of the Three Set of Dynamic Images on the Validation Set of the ChaLearn LAP IsoGD Dataset.  RP Denotes Conventional Rank Pooling; HRP Represents Hierarchical Rank Pooling.\label{comlimentary}}
\begin{tabular}{|c|c|c|}
\hline
Method & Accuracy for RP & Accuracy for HRP\\
\hline
DDI (forward) & 36.13\% & 36.92\%\\
\hline
DDI (backward) & 30.45\% & 31.24\%\\
\hline
DDI (fusion) & 37.52\% & 37.68\% \\
\hline
DDNI (forward) & 24.86\% & 25.02\%\\
\hline
DDNI (backward) &  24.58\% & 24.64\%\\
\hline
DDNI (fusion)& 29.26\% & 29.48\%\\
\hline
DDMNI (forward) & 24.81\% & 24.69\%\\
\hline
DDMNI (backward) & 23.14\% & 23.57\%\\
\hline
DDMNI (fusion) & 27.75\% & 27.89\%\\
\hline
Fusion All & 42.56\% & 43.72\% \\
\hline
\end{tabular}
\end{table}

The results obtained by the proposed method on the validation and test sets are listed and compared with previous methods in Table~\ref{table2}. These methods include MFSK combined 3D SMoSIFT~\cite{wan20143d} with (HOG, HOF and MBH)~\cite{wang2013action} descriptors.  MFSK+DeepID further 
included Deep hidden IDentity (Deep ID) feature~\cite{sun2014deep}. Thus, these 
two methods utilized not only hand-crafted features but also deep learning 
features. Moreover, they extracted features from RGB and depth separately, 
concatenated them together, and adopted Bag-of-Words (BoW) model as the final 
video representation. The other methods, 
WHDMM+SDI~\cite{pichaoTHMS,bilen2016dynamic}, extracted features and conducted 
classification with ConvNets from depth and RGB individually and adopted 
multiply score fusion for final recognition. SFAM~\cite{Pichaocvpr2017} adopted 
scene flow to extract features and encoded the flow vectors into action maps, 
which fused RGB and depth data from the onset of the process. C3D~\cite{yunanli} applied 3D convolutional networks  to both depth and RGB channels and fused them in a late fusion method. Pyramidal 3D CNN~\cite{guangming} adopted 3D convolutional networks to pyramid input to recognize gesture from both clip videos and entire video. It is noteworthy that the results of the proposed method have been obtained using a single modality viz., depth data, while all compared methods are based on RGB and depth modalities. From this table, we can see that the proposed method outperformed all of these recent works significantly, and illustrated its effectiveness.

\begin{table}[!ht]
\centering
\caption{Comparative Accuracy of Proposed Method and Baseline 
Methods on the ChaLearn LAP IsoGD Dataset.  \label{table2}}
\begin{tabular}{|c|c|c|}
\hline
Method & Set & Recognition rate $r$\\
\hline
MFSK~\cite{pami16Jun} & Validation & 18.65\%\\
\hline
MFSK+DeepID~\cite{pami16Jun} & Validation & 18.23\%\\
\hline
SDI~\cite{bilen2016dynamic} & Validation & 20.83\%\\
\hline
WHDMM~\cite{pichaoTHMS} & Validation & 25.10\%\\
\hline
Scene Flow~\cite{Pichaocvpr2017} & Validation & 36.27\%\\
\hline
Proposed Method & Validation & 43.72\%\\
\hline
MFSK~\cite{pami16Jun} & Testing & 24.19\% \\
\hline
MFSK+DeepID~\cite{pami16Jun} & Testing & 23.67\%\\
\hline
Pyramidal 3D CNN~\cite{guangming}  & Testing & 50.93\%\\
\hline
C3D~\cite{yunanli} & Testing & 56.90\%\\
\hline
Proposed Method & Testing &59.21\%\\
\hline
\end{tabular}
\end{table}

%
 
 \begin{table}[!ht]
\centering
\caption{Accuracies of the Proposed Method and Previous 
Methods on the ChaLearn LAP ConGD Dataset. \label{table3}}
\begin{tabular}{|c|c|c|}
\hline
Method & Set & Mean Jaccard Index $\overline{J_{S}}$\\
\hline
MFSK~\cite{pami16Jun} & Validation & 0.0918\\
\hline
MFSK+DeepID~\cite{pami16Jun} & Validation & 0.0902\\
\hline
Proposed Method & Validation & 0.3905\\
\hline
MFSK~\cite{pami16Jun} & Testing & 0.1464 \\
\hline
MFSK+DeepID~\cite{pami16Jun} & Testing & 0.1435\\
\hline
IDMM + ConvNet~\cite{pichaoicprwa} & Testing & 0.2655\\
\hline
C3D~\cite{Necati} & Testing & 0.2692\\
\hline
Two-stream RNNs~\cite{xiujuan} & Testing & 0.2869\\
\hline
Proposed Method & Testing &0.4109\\
\hline
\end{tabular}
\end{table}

\begin{figure}[t]
\begin{center}
{\includegraphics[height = 40mm, width = 75mm]{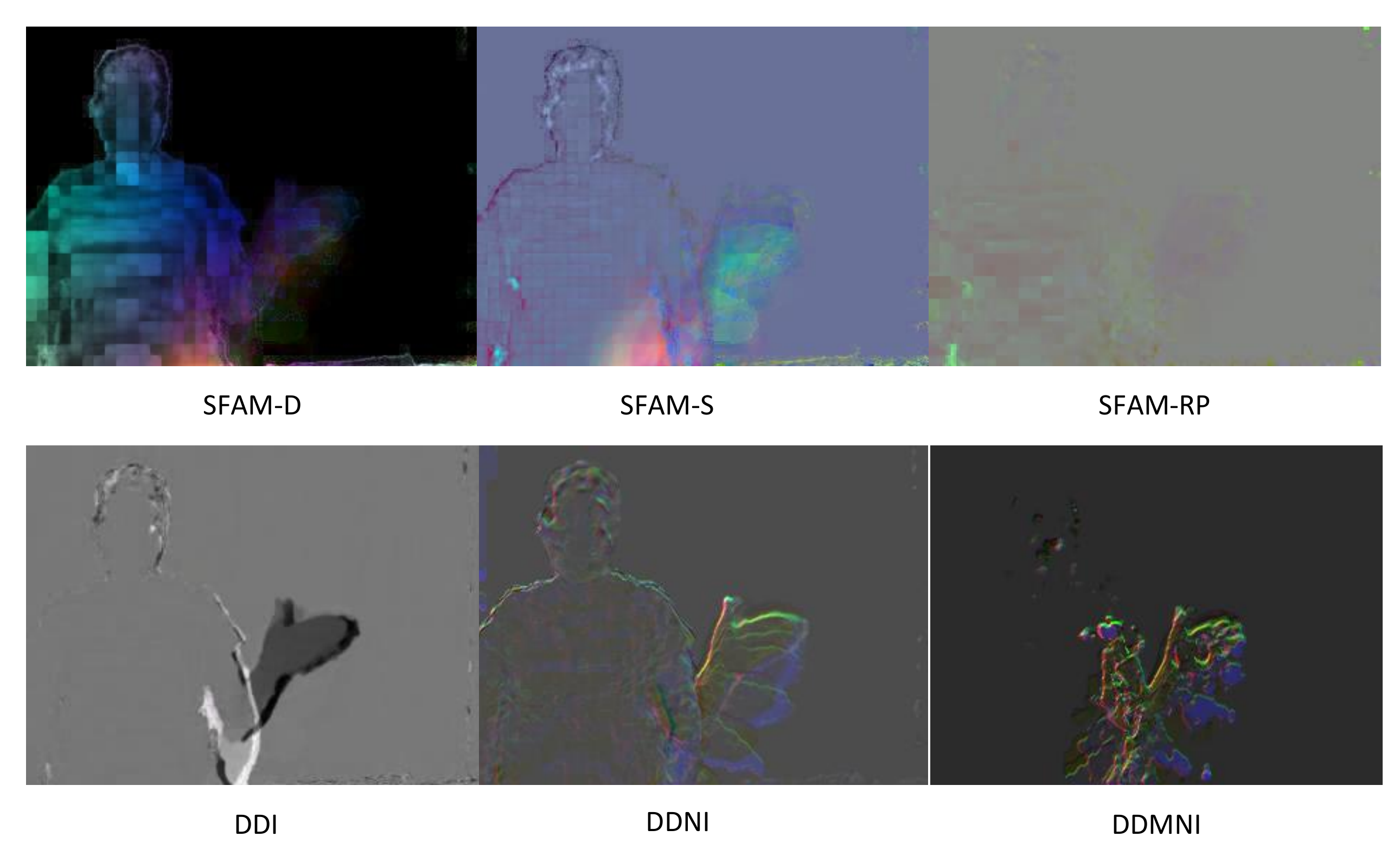}}
\end{center}
\caption{Visual comparion of the DDIs in the proposed method with SFAM~\cite{Pichaocvpr2017}.}
\label{compare}
\end{figure}

Fig~\ref{compare} visually compares the DDI, DDNI and DDMI and the SFAM-D, SFAM-S and SFAM-RP proposed in~\cite{Pichaocvpr2017}. It can be seen that the proposed DDIs do not suffer the blocky artifacts as the SFAMs do and the motion information in DDIs becomes much clean.

\subsection{ChaLearn LAP ConGD Dataset}
\subsubsection{Description}
The ChaLearn LAP ConGD Dataset is also derived from the ChaLearn Gesture Dataset (CGD)~\cite{guyon2014chalearn}. It has 47933 RGB-D  gesture instances in 22535 RGB-D gesture videos. Each RGB-D video may contain one or more gestures.  There are 249 gestures performed by 21 different individuals.  The detailed information of this dataset is shown in Table~\ref{Con}. In this paper, only depth data was used in the proposed method. 

\begin{table*}[!ht]
\centering
\caption{Information of the ChaLearn LAP ConGD Dataset. \label{Con}}
\begin{tabular}{|c|c|c|c|c|c|c|c|}
\hline
Sets &\# of labels &\# of gestures & \# of RGB videos & \# of depth videos & \# of subjects & label provided & temporal segment provided\\
\hline
Training & 249 & 30442 & 14134 & 14134 & 17 & Yes & Yes\\
\hline
Validation & 249 & 8889 & 4179 & 4179 & 2 & No & No\\
\hline
Testing & 249 &  8602 & 4042 & 4042 & 2 & No & No\\
\hline
All & 249 & 47933 & 22535 & 22535 & 21 & - & -\\
\hline
\end{tabular}
\end{table*}


\subsubsection{Evaluation Protocol}

The dataset is divided into training, validation and test sets by the challenge organizers. All three sets include data from different subjects and the gestures of one subject in validation and test sets do not appear in the training set. Jaccard index (the higher the better) is adopted to measure the performance. The Jaccard index measures the average relative overlap between true and predicted sequences of frames for a given gesture. For a sequence $s$, let $G_{s,i}$ and $P_{s,i}$ be binary indicator vectors for which 1-values correspond to frames in which the $i^{th}$ gesture label is being performed. The Jaccard Index for the $i^{th}$ class is defined for the sequence $s$ as:

\begin{equation}
J_{s,i} = \dfrac{G_{s,i}\bigcap P_{s,i}}{G_{s,i}\bigcup P_{s,i}},
\end{equation}
where $G_{s,i}$ is the ground truth of the $i^{th}$ gesture label in sequence $s$, and $P_{s,i}$ is the prediction for the $i^{th}$ label in sequence $s$. When $G_{s,i}$ and $P_{s,i}$ are empty, $J_{(s,i)}$ is defined to be 0. Then for the sequence $s$ with $l_{s}$ true labels, the Jaccard Index $J_{s}$ is calculated as:

\begin{equation}
J_{s} = \dfrac{1}{l_{s}}\sum_{i=1}^{L}J_{s,i}.
\end{equation}
For all testing sequences $S = {s_{1},...,s_{n}}$ with $n$ gestures, the mean Jaccard Index $\overline{J_{S}}$ is used as the evaluation criteria and calculated as:
 \begin{equation}
 \overline{J_{S}} = \dfrac{1}{n}\sum_{j=1}^{n}J_{s_{j}}.
 \end{equation}
 
  \begin{table*}[!ht]\small
  \centering
  \caption{Comparative Accuracy of the Three Set of Dynamic Images on the NTU RGB+D Dataset. RP Denotes Conventional Rank Pooling; HRP Represents Hierarchical Rank Pooling.\label{tableNTUcom} }
\begin{tabular}{|c|c|c|c|c|}
  \hline
  Method&	\tabincell{c}{Cross subject\\ Accuracy for RP} &	\tabincell{c}{Cross subject\\ Accuracy for HRP} &  \tabincell{c}{Cross view\\ Accuracy for RP}	&  \tabincell{c}{Cross view\\ Accuracy for HRP}\\
    \hline
DDI (forward)	&75.80\%	 & 76.10\% & 76.50\%& 76.75\%\\
 \hline
DDI (backward)	&70.99\%	 & 75.45\% & 75.62\% & 75.48\%\\
 \hline
DDI (fusion) 	&81.66\% &	82.01\% & 81.53\%& 81.60\%\\
 \hline
DDNI (forward)	&79.79\%	&79.98\% & 54.57\%& 55.01\%\\
 \hline
DDNI (backward) & 81.46\% & 81.28\%& 56.61\% & 57.43\%\\
 \hline
DDNI (fusion) & 84.18\%	& 84.24\%& 61.07\%&	62.35\%\\
 \hline
DDMNI (forward) & 68.89\% & 69.33\% & 50.01\% & 50.67\%\\
 \hline
DDMNI (backward) & 70.04\% 	& 71.11\% & 49.53\% & 49.27\%\\
 \hline
DDMNI (fusion) &	 73.56\%	& 74.27\% & 54.98\% & 55.09\%\\
 \hline
Fusion All &	 86.72\% & 87.08\% & 83.75\% & 84.22\%\\
\hline
\end{tabular}
\end{table*}
 
\subsubsection{Experimental Results}

To measure the performance of the segmention, the Levenshtein distance as used in~\cite{Lea2016} between the segmented sequence and the ground truth segments is caculated as the metric. The QOM method have achived on average score of $71.44$ on the ChaLearn LAP ConGD Dataset, which is higher than what the best result, i.e. $66.56$ in~\cite{Lea2016}, on a similar dataset probably due to the clear transitional pose between two gestures in the ChaLearn LAP ConGD Dataset.

The recognition results of the proposed method on the validation and test sets and their comparisons with the results of previous methods are shown in Table~\ref{table3}. MFSK and MFSK+DeepID~\cite{pami16Jun} methods first segmented the continuous videos to segments and then extracted the features over the segments over two modalities to train and classify the actions. IDMM + ConvNet~\cite{pichaoicprwa} also adopted the action segmentation method and then extracted one improved depth motion map using color coding method over the segments, and ConvNet was adopted to train and classify segmented actions. C3D~\cite{Necati} applied 3D convolutional networks to RGB video and jointly learn the features  and classifier. Two-stream RNNs~\cite{xiujuan} first adopted R-CNN to extract the hand and then conducted temporal segmentation. Two-stream RNNs were adopted to fuse  multi-modality features for final recognition based on segments. The results showed that the proposed method outperformed all previous methods largely, even though only single modality, i.e. depth data, was used.


\subsection{ NTU RGB+D Dataset}
\subsubsection{Description}
 To our best knowledge, NTU RGB+D Dataset is currently the largest action 
recognition dataset in terms of training samples for each action. The 3D data is 
captured by Kinect v2 cameras. The dataset has more than 56 thousand sequences 
and 4 million frames, containing 60 actions performed by 40 subjects aged 
between 10 and 35. It consists of front view, two side views and left, right 45 
degree views. This dataset is challenging due to large intra-class and 
viewpoint variations. 
\subsubsection{Evaluation Protocol}
For fair comparison and evaluation, the same protocol as 
that in~\cite{shahroudy2016ntu} was used. It has both cross-subject and 
cross-view evaluation. In the cross-subject evaluation, samples of subjects 1, 
2, 4, 5, 8, 
9, 13, 14, 15, 16, 17, 18, 19, 25, 27, 28, 31, 34, 35 and 38 were used as 
training and samples of the remaining subjects were reserved for testing. In the 
cross-view evaluation, samples taken by cameras 2 and 3 were used as training, 
while the testing set includes samples from camera 1.

\subsubsection{Experimental Results}
 Similarly to LAP IsoGD Dataset, we conducted several 
experiments to compare the three set of dynamic images using conventional rank pooling method and the proposed hierarchical bidirectional rank pooling method. The comparisons are shown in Table~\ref{tableNTUcom}. From the Table it can be seen that compared with DDIs, DDNIs achieved much better results than DDI in cross-subject setting, due to the sensitivity of norm vectors to motion over real depth values. This justified the effectiveness of proposed depth norm images for rank pooling.  However, due to the sensitivity of norm vectors to motion and view angles, in cross-view setting, much worse results were achieved for DDNIs and DDMNIs.  From the final fusion results we can see that the three set of dynamic images exploit the shape and motion at different levels, and  provide complimentary information to each other. 

Table~\ref{tableNTU} lists the performance of the proposed method and those 
previous works. The proposed method was compared with some skeleton-based methods and 
depth-based methods previously reported on this dataset. We can see that the 
proposed method outperformed all the previous works significantly. 

 \begin{table}[h]\small
  \centering
 \caption{Accuracies of the Proposed Method and Previous Methods on NTU RGB+D Dataset.\label{tableNTU} }
 \begin{tabular}{|c|c|c|c|}
  \hline
  Method&	Modality & Cross	&  Cross\\
& & Subject & View \\
    \hline
Lie Group~\cite{vemulapalli2014human}	& Skeleton &50.08\%	 & 52.76\%\\

HBRNN~\cite{du2015hierarchical}	& Skeleton &59.07\% 	 & 63.97\%\\

2 Layer RNN~\cite{shahroudy2016ntu}	& Skeleton &56.29\%  &	64.09\%\\

2 Layer LSTM~\cite{shahroudy2016ntu}	& Skeleton &60.69\%	&67.29\%\\

Part-aware LSTM~\cite{shahroudy2016ntu}& Skeleton & 	62.93\%	&70.27\%\\

ST-LSTM~\cite{liu2016spatio} & Skeleton &65.20\%	&76.10\%	\\
ST-LSTM+ Trust Gate~\cite{liu2016spatio} & Skeleton &	69.20\%	&77.70\%\\
JTM~\cite{pichao2016} & Skeleton & 73.40\% & 75.20\%\\
\hline
HON4D~\cite{Oreifej2013} & Depth & 30.56\% & 7.26\%\\
SNV~\cite{yangsuper} & Depth & 31.82\% & 13.61\%\\
SLTEP~\cite{ji2017spatial}& Depth & 58 .22\% & -- \\
\hline
Proposed Method	& Depth &87.08\%&	84.22\%\\
\hline
\end{tabular}
\end{table}

\begin{figure}[ht]
\begin{center}
{\includegraphics[height = 75mm, width = 90mm]{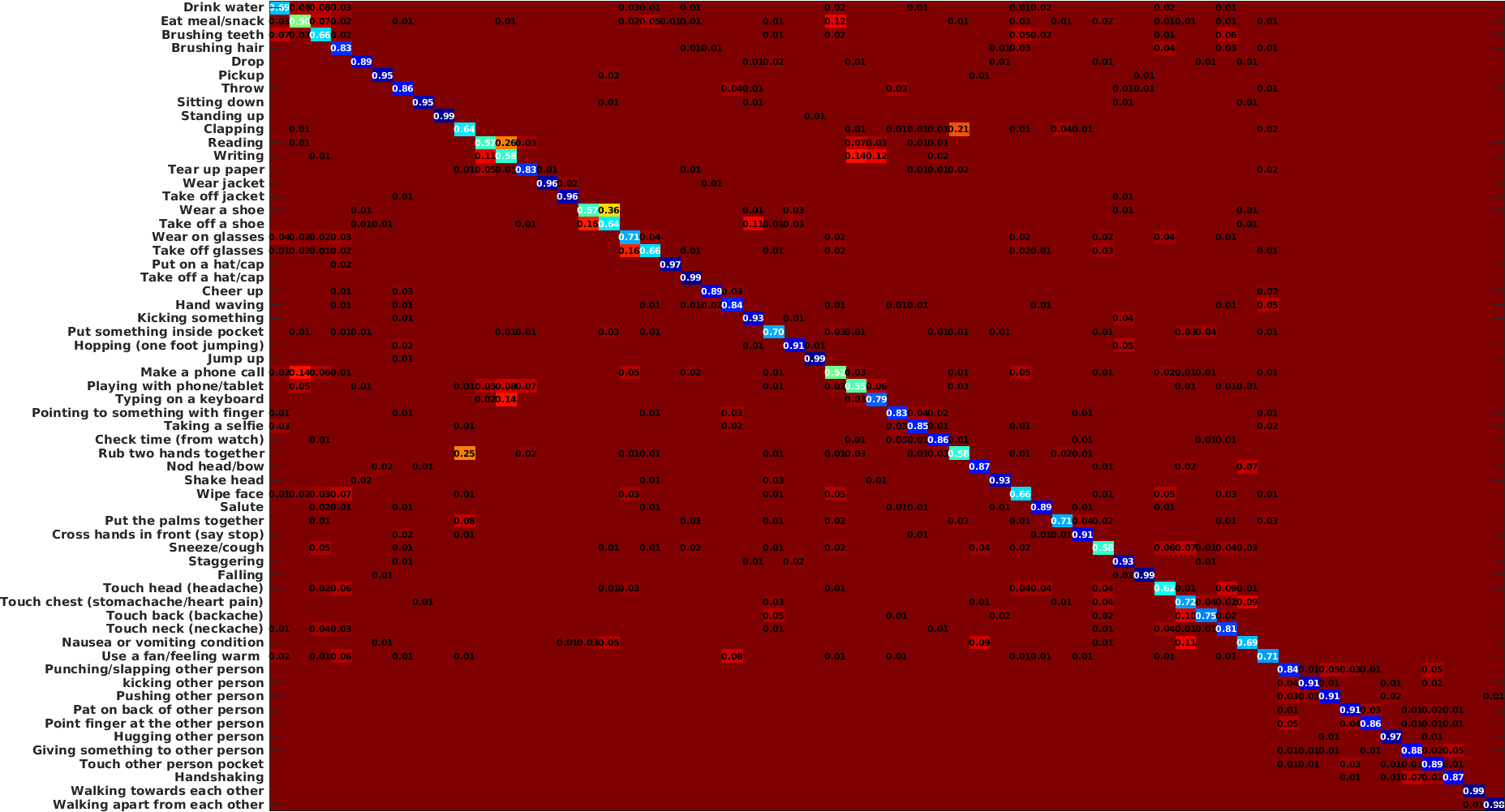}}
\end{center}
\caption{The confusion matrix for the cross subject setting on NTU RGB+D dataset.}
\label{cm}
\end{figure}

The confusion matrix of the cross subject setting is shown in Fig~\ref{cm}. We can see that the method distinguishes well the single subject based actions from the two subject based actions. Human-object interactions like ``Make a phone call" may be confused with other actions with similar motion patterns such as ``Eat meal/snack" and ``brush teeth", because the objects that involved in the actions are hard to distinguish in the depth maps, which is the weakness of using depth modality alone. Other single subject based actions such as ``Rub two hands together" and ``Clapping" are also easily confused with each other due to the similarity of the two actions in motion patterns.

\subsection{Discussions}
\subsubsection{Cross-dataset analysis}
It is not feasible to train the model on a dataset such as NTU RGB+D and directly test on another dataset such as ChaLearn LAP IsoGD because of the different sets of actions in the two datasets. However, we conducted the following experiments.  The model trained over the DDI forward channel on NTU RGB+D dataset was adopted as pre-trained model and fine-tuned on the ChaLearn LAP IsoGD dataset and vice verse. No noticeable improvement on the results have been demonstrated as shown in Table~\ref{cs}. It is probably because both datasets are large enough to train a good model over the pre-trained model obtained from ImageNet.
\begin{table}[!ht]
\centering
\caption{Cross-dataset training.\label{cs}}
\begin{tabular}{|c|c|}
\hline
Setting & Accuracy \\
\hline
Fine-tune on ChaLearn LAP IsoGD from ImageNet model & 36.92\%\\
\hline
Fine-tune on NTU RGB+D  from ImageNet model & 76.10\%\\
\hline
\tabincell{c}{Pre-train on NTU RGB+D dataset(cross-subject setting)\\Fine-tune on ChaLearn LAP IsoGD dataset} & 36.72\% \\
\hline
\tabincell{c}{Pre-train on ChaLearn LAP IsoGD dataset\\Fine-tune on NTU RGB+D dataset (cross-subject setting)} & 75.95\% \\
\hline
\end{tabular}
\end{table}

\subsubsection{Different fusion methods}
In this paper, we adopt product score fusion to fuse the classifications obtained from the six channels. The other two commonly used late score fusion methods are average and maximum score fusion. The comparisons among the three late score fusion methods are shown in Table~\ref{fusion}. We can see that the product score fusion method achieved the best results on all the three large datasets. This verifies to a large extent that the six channels are likely to be statistically independent and product score fusion is the right choice. We  also extracted the convnet features (the last fc 4096 feature) from the six models over ChaLearn LAP Iso dataset and concatenate the six feature vectors into one long feature vector (4096*6) and using linear SVM to do clissification. 
The result on the validation  set is 40.88\% compared with $43.73\%$  achieved by the proposed method.

\begin{table}[!th]
\centering
\caption{Comparison of three different late score fusion methods on the six datasets.\label{fusion}}
\begin{tabular}{|c|c|c|c|} \hline
\multirow{3}{*}{Dataset}
& \multicolumn{3}{c|}{Score Fusion Method}\\
 \cline{2-4}
 & Max  & Average  & Product \\ \hline
  ChaLearn LAP IsoGD & 40.38\%  & 42.21\% & 43.72\%\\ \hline
  ChaLearn LAP ConGD & 0.3591  & 0.3768 & 0.3905\\ \hline
  NTU RGB+D (cross subject) & 84.62\%  & 85.77\% & 87.08\%\\ \hline
  NTU RGB+D (cross view) & 82.47\%  & 83.56\% & 84.22\%\\ \hline
\end{tabular}
\end{table}

\subsubsection{Channel fusion}
We conducted several experiments over ChaLearn LAP Iso dataset by using two and three path fusion to analysis the channel fusion cases, and the  results are shown in Table~\ref{channelfusion}.
  \begin{table}[!ht]
\centering
\caption{Comparative results for two paths fusion over ChaLearn LAP Iso  validation  dataset.\label{channelfusion}}
\begin{tabular}{|c|c|}
\hline
Fusion Channels &  Accuracy \\
\hline
DDI+DDNI & 41.45\%\\
\hline
DDI+DDMNI & 39.12\%\\
\hline
DDNI+DDMNI& 33.64\%\\
\hline
DDI+DDNI+DDMNI&43.72\%\\
\hline
\end{tabular}
\end{table}
From the Table we can see that the fusion of DDI with either DDNI or DDMNI achieved better results than the fusion of DDNI and DDMNI, and the fusion of the three representation has led to the best results.
\\

\subsubsection{Computational cost}
we compared the computational cost of the proposed method in the test phase without any code optimization with the computational cost of the RGB+D-based methods MFSK+DeepID~\cite{pami16Jun}, SFAM~\cite{Pichaocvpr2017}, depth-based method WHDMM~\cite{pichaoTHMS} on ChaLearn LAP IsoGD dataset. The results are reported as follows
 \begin{table}[!ht]
\centering
\caption{Ccomputational cost comparison of the proposed method (without any code optimization) with previous methods on the ChaLean LAP IsoGD dataset. CPU time refers the computation of representation (DDI/DDNI/DDMNI) and the GPU time represents the time taken by the convnet feature extraction. For each path in the proposed method the computatinal time is roughly same due to the fast preprocessing module.\label{cc}}
\begin{tabular}{|c|c|}
\hline
Method & Computational Cost \\
\hline
MFSK+DeepID~\cite{pami16Jun} & 41.00s(CPU)++00.00ms(GPU)\\
\hline
SFAM~\cite{Pichaocvpr2017} & 6.30s(CPU)+31.00ms(GPU)\\
\hline
WHDMM~\cite{pichaoTHMS} & 0.60s(CPU time)+16.00ms(GPU time) \\
\hline
Proposed Method & 62.00s(CPU)+31.00ms(GPU time)\\
\hline
\end{tabular}
\end{table}
Though the proposed method requires more computation compared with the previous methods, the significant improvement on performance would justify the additional computation is worthwhile.

\section{Conclusions}
\label{sec:conclusions}
To fully exploit the spatial, temporal and structural information in a depth sequence for action recognition at different time-scale, this paper presents three simple, compact yet effective representations. These representations are constructed through the extended ranking pooling, namely, the hierarchical bidirectional rank pooling. The representations enable fine-tuning on depth data existing ConvNets models that are learned from RGB video. Experimental results on three large datasets ChaLearn LAP IsoGD, ChaLearn LAP ConGD and NTU RGB+D have verified the efficacy of the proposed representations.

\section*{Acknowledgment}
The authors would like to thank NVIDIA Corporation for the donation of a Tesla K40 GPU card and a Pascal TITAN X card used in this research. The authors would also like to thank Dr. Jun Wan for validating the final results on the first two datasets.

\bibliographystyle{IEEEtran}




\begin{IEEEbiography}[{\includegraphics[width=1in,height=1.25in,clip,keepaspectratio]{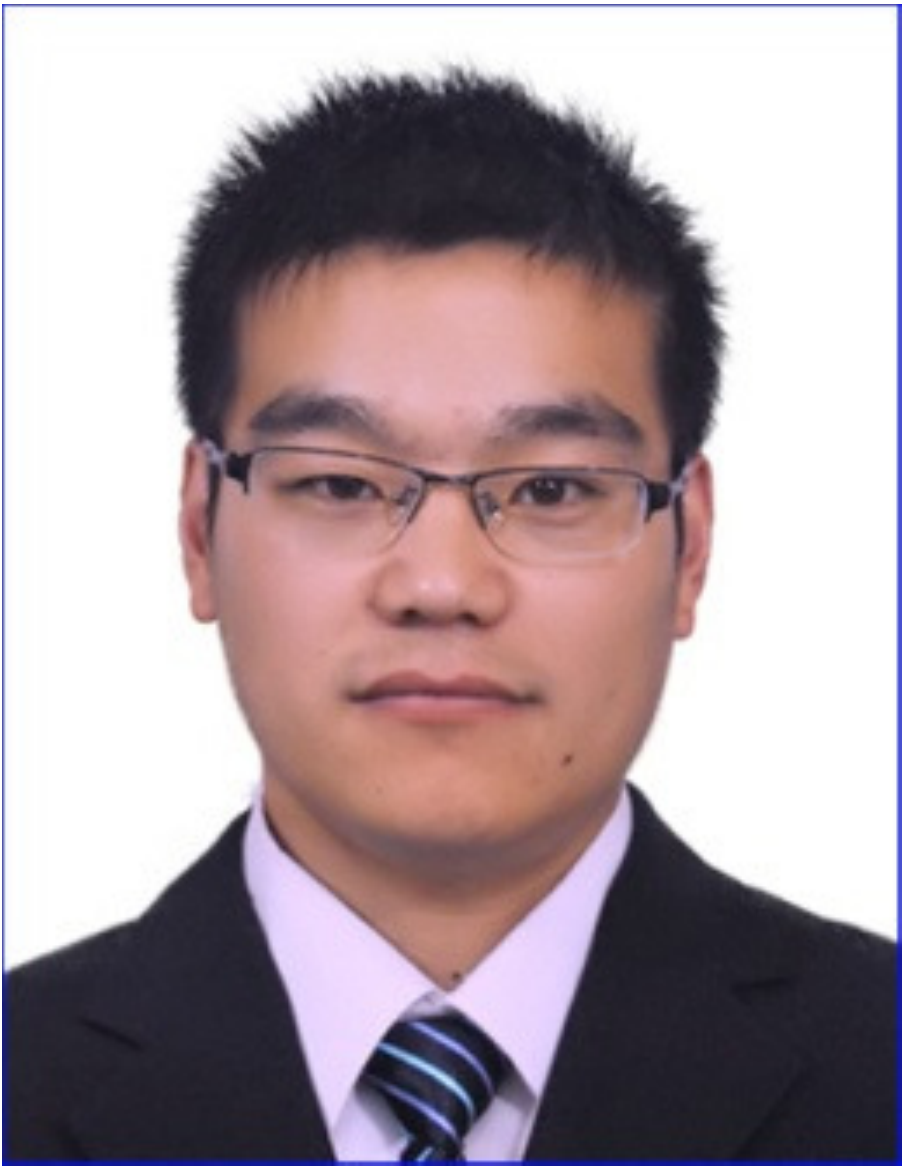}}]{Pichao Wang}

(M'17) received his PhD in computer science from The University of Wollongong, Australia. He was a recipient of the 2016 ChaLearn Isolated Gesture Recognition \& Continuous Gesture Recognition Challenge Awards, at ICPR 2016. He was also a recipient of two Best Paper Awards at the ChaLearn Look at People Challenge, 2016. He has published 30+ peerreviewed papers, including those in highly regarded journals and conferences such as IEEE TMM, IEEE THMS, CVPR, AAAI, ACM MM, etc. He served on the Technical Program Committees of ICCV 2017, CVPR 2018, ICME 2018, and IJCAI 2018.
His current research 
interests include computer vision and machine learning. He is now a researcher in Motovis Inc. He is a member of IEEE.
\end{IEEEbiography}
\vspace{-1cm}
\begin{IEEEbiography}[{\includegraphics[width=1in,height=1.25in,clip,keepaspectratio]{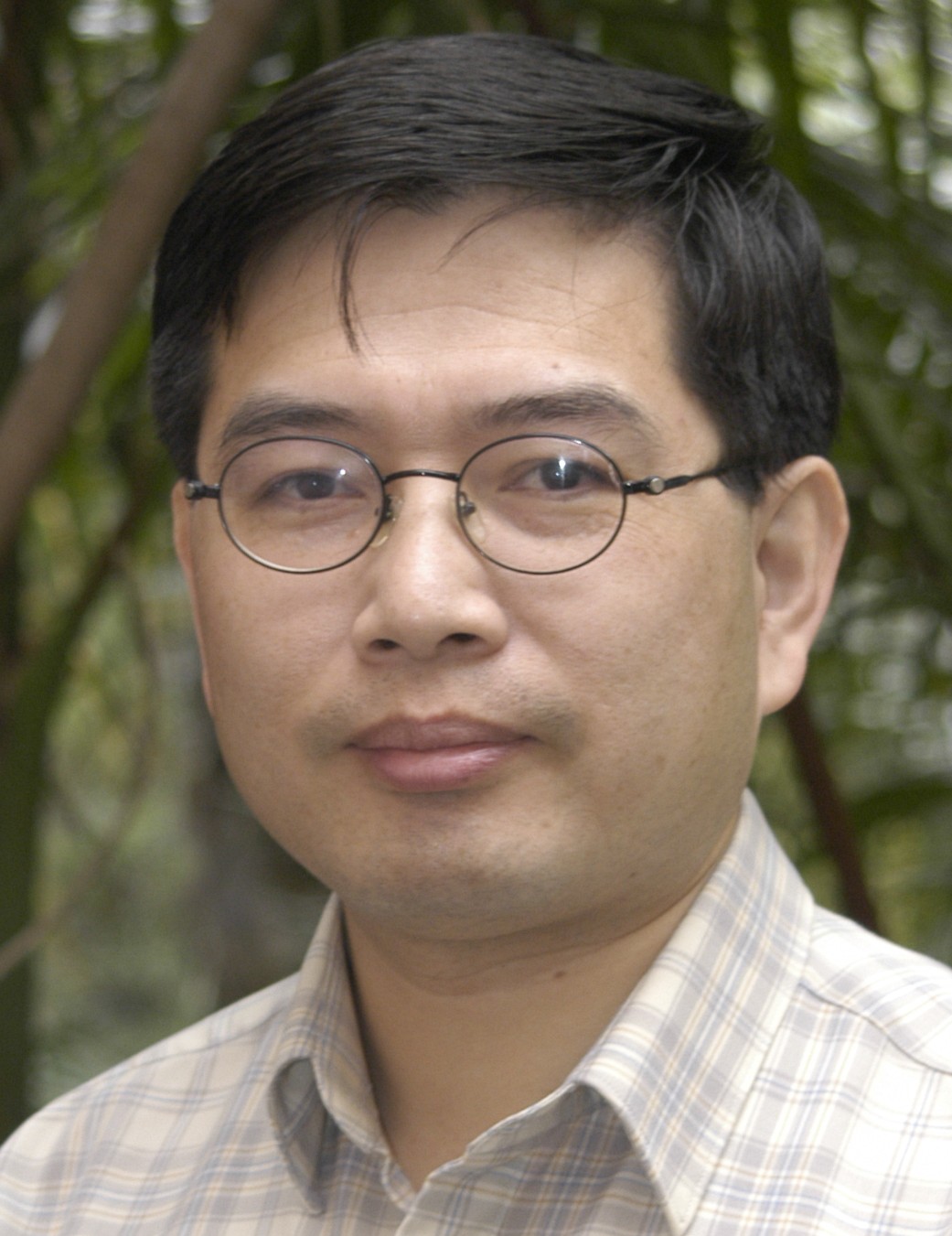}}]{Wanqing Li}
(SM'12) received his PhD in electronic engineering from The University of Western Australia. He was an Associate Professor (91-92) at Zhejiang University, a Senior Researcher and later a Principal Researcher (98-03) at Motorola Research Lab, and a visiting researcher (08,10 and 13) at Microsoft Research US. He is currently an Associate Professor and Co-Director of Advanced Multimedia Research Lab (AMRL) of University of Wollongong, Australia.  His research areas are 3D computer vision, 3D multimedia signal processing and medical image analysis. Dr. Li is a Senior Member of IEEE.

\end{IEEEbiography}

\vspace{-1cm}
\begin{IEEEbiography}[{\includegraphics[width=1in,height=1.25in,clip,keepaspectratio]{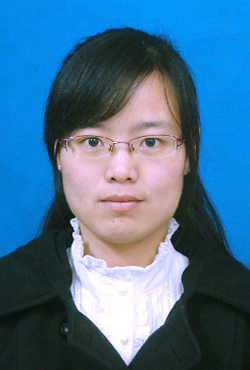}}]{Zhimin Gao}

received the BS degree from North China Electric Power University, 
China in 2010,  the MS degree at Tianjin University, China in 2013 and the PhD degree
 with the School of Computing and Information 
Technology, University of Wollongong, Australia in 2018.  Her research 
interests include Computer Vision and Machine Learning.
\end{IEEEbiography}
\vspace{-1cm}

\begin{IEEEbiography}[{\includegraphics[width=1in,height=1.25in,clip,keepaspectratio]{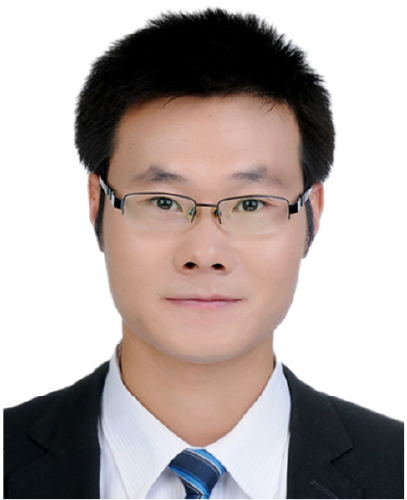}}]{Chang Tang}

(S'13) received his Ph.D. degree from Tianjin University, Tianjin, China in 2016. He joined the AMRL Lab of the University of Wollongong between Sep. 2014 and Sep. 2015. He is now an associate professor at the School of Computer Science, China University of Geosciences, Wuhan, China. His current research interests include machine learning and data mining.

\end{IEEEbiography}

%

\vspace{-1cm}
\begin{IEEEbiography}[{\includegraphics[width=1in,height=1.25in,clip,
keepaspectratio]{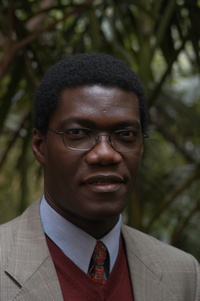}}]{Philip O. Ogunbona}

(SM'97) earned his PhD, DIC (Electrical 
Engineering) from Imperial College, London. His 
Bachelors degree was in Electronic and Electrical 
Engineering from the University of Ife, Nigeria 
(now named Obafemi Awolowo University). 
He is a Professor and Co-Director of the 
Advanced Multimedia Research Lab, University of Wollongong, Australia. His 
research interests include computer vision, pattern recognition and machine 
learning. He is a Senior Member of IEEE and Fellow of Australian Computer Society.
\end{IEEEbiography}

\end{document}